\documentclass{article}
\usepackage{tencent_ailab_tech_report}
\usepackage[colorlinks = true,
            linkcolor = blue,
            urlcolor  = blue,
            citecolor = blue,
            anchorcolor = blue]{hyperref}
\usepackage{microtype}
\usepackage{hyperref}
\usepackage{xurl}
\usepackage{booktabs}
\usepackage{enumitem}
\usepackage{multicol}
\usepackage{CJKutf8}
\usepackage{amsmath}
\usepackage{siunitx}
\usepackage{floatflt}
\usepackage{graphicx}
\usepackage{booktabs}
\usepackage{wrapfig}
\usepackage{authblk}

\usepackage[ruled,vlined]{algorithm2e}  
\usepackage{mathrsfs,algorithmic}

\usepackage{microtype}
\usepackage{subfig}
\usepackage{multirow}
\usepackage{booktabs}   
\usepackage{pifont}     
\usepackage{hyperref}
\usepackage{amssymb}

\usepackage{tabularx}
\usepackage{booktabs}       
\usepackage{nicefrac}       
\usepackage{microtype}      
\usepackage{xcolor}         
\usepackage{bbm, dsfont}
\usepackage{array}
\usepackage{amsthm}
\usepackage{bm}
\usepackage{stfloats}
\usepackage{mathtools}

\usepackage{array}
\usepackage{amsmath}
\usepackage{amssymb}
\usepackage{mathtools}
\usepackage{amsthm}

\usepackage[capitalize,noabbrev]{cleveref}
\usepackage{adjustbox}

\theoremstyle{plain}

\theoremstyle{definition}

\theoremstyle{remark}

\usepackage{xcolor}

\colmfinalcopy

\usepackage{soul}
\usepackage{fontawesome5}
\usepackage{wrapfig}
\usepackage{caption}
\usepackage{colortbl}
\usepackage{mathabx}
\usepackage{arydshln}

\definecolor{nred}{RGB}{196, 38, 11}
\definecolor{ngreen}{RGB}{18, 141, 21}
\definecolor{nblue}{RGB}{41, 52, 190}
\definecolor{hzw}{RGB}{223, 97, 76}
\definecolor{lt}{RGB}{54, 89, 170}
\definecolor{tblue}{rgb}{0.867, 0.922, 0.969}
\definecolor{zlblue}{RGB}{196, 223, 251}
\definecolor{exp_table_blue}{HTML}{DAECED}
\definecolor{mygray}{gray}{0.92}

\newcommand{\ignore}[1]{}

\hypersetup{
    colorlinks=true,
    linkcolor=purple,
    citecolor=nblue,
    filecolor=magenta,
    urlcolor=nblue
}

\crefname{section}{\S}{\S\S}
\Crefname{section}{\S}{\S\S}
\crefname{appendix}{appendix}{appendix}

\title{$\\ \\$Dynamic Important Example Mining for Reinforcement Finetuning}

\author{%
Haoru Tan$^{1,2}\thanks{Personal Email: \textit{hrtan@eee.hku.hk}}$
~~~Sitong Wu$^{3,\dagger}$
~~~Yanfeng Chen$^{2}$
~~~Shizhen Zhao$^{1}$
~~~Yang-Tian Sun$^{1}$
~~~Tianjia Liu$^{1}$ 
~~~Chirui Chang$^{1}$
~~~Shaofeng Zhang
~~~Samm Sun$^{2}$
~~~Xiuzhe Wu$^{4}$
~~~Ruobing Xie$^{2}$
~~~Xiaojuan Qi$^{1,\dagger}$ \\
\vspace{10pt}
$^1$HKU \ \ \ \ \ \ $^2$Hunyuan Team, Tencent \ \ \ \ \ \ $^3$CUHK \ \ \ \ \ \ $^4$Stanford \ \ \\
}

\colmfinalcopy 

\begin{document}

\maketitle

\renewcommand\thefootnote{\fnsymbol{footnote}}
\footnotetext[2]{$\dagger$ Corresponding author: Sitong Wu and Xiaojuan Qi.}
\renewcommand\thefootnote{\arabic{footnote}}

\begin{abstract}

Reinforcement fine-tuning (RFT) is increasingly used to strengthen the reasoning abilities of large models, yet its effectiveness is bound by how training data are selected and used. Most data-centric RFT methods rely on static or heuristic sample selection, implicitly assuming a sample's value is fixed over training. This overlooks the non-stationary dynamics of policy learning and can lead to suboptimal updates.
We propose \textbf{Dynamic Important Example Mining} (\textbf{DIEM}), a principled and fully automated framework that makes data utilization adaptive throughout RFT. DIEM integrates two components into each optimization step: (i) a gradient-alignment importance estimator that efficiently approximates each sample's marginal contribution to policy improvement; and (ii) a constrained batch reweighting scheme that maximizes aggregate utility while preserving the update's gradient magnitude to stabilize optimization.
Across several reasoning benchmarks, DIEM consistently outperforms strong static and dynamic baselines. The code will be released via https://github.com/hrtan/DIEM.

\end{abstract}

\section{Introduction}
\label{sec:introduction}

Reinforcement Fine-Tuning (RFT) has recently emerged as a key paradigm for enhancing the reasoning and alignment capabilities of large-scale vision--language models \citep{tong2024cambrian1,llava-rasoner,bai2025qwen25vltechnicalreport,chen2024internvl,zhu2025internvl3,wang2025internvl3_5,li2024llava} and language models \citep{deepseek-r1, o1, team2025kimi, open-reasoner-zero}. By integrating reinforcement learning into the post-training stage, RFT allows models to learn directly from reward signals, refining their decision-making beyond supervised imitation \citep{chu2025sft}. However, the effectiveness of RFT critically depends on how data is utilized during policy optimization \citep{deepseek-r1, li2025limr, deepscaler2025}. The selection, ordering, and weighting of training samples fundamentally determine both the stability of optimization and the generalization of reasoning abilities.

Despite RFT's inherently dynamic, non-stationary nature, most existing data-centric approaches overlook this dynamism, treating data importance as fixed throughout training. These methods typically pre-select samples using heuristic indicators such as reward variance \citep{wang20251datareinforcement} or difficulty \citep{li2025limr}. While such strategies can improve data efficiency, they suffer from a critical limitation: the assumption that a sample's contribution remains constant across the entire learning trajectory.

Recent advances in reinforcement learning \citep{gao2025pcl, zhang2025speedrlfastertrainingreasoning, yu2025dapo} have begun to explore dynamic sample weighting and curriculum scheduling within the RFT framework. For example, Prompt Curriculum Learning (PCL) \citep{gao2025pcl} introduces an auxiliary value model to estimate sample difficulty, while SPEED-RL \citep{zhang2025speedrlfastertrainingreasoning} prioritizes data with intermediate pass rates to maintain balanced learning progression. However, such heuristic metrics (e.g., ``difficulty score'') fail to capture two essential aspects.
First, they do not accurately reflect the policy's intrinsic preference or fitness for a given sample, since the judgment is made externally, detached from the policy itself.
Second, they cannot quantify a sample's true marginal impact on policy updates, which is crucial for understanding and optimizing the contribution of each example to reasoning improvement.

To address these limitations, we propose \textbf{Dynamic Important Example Mining} (\textbf{DIEM}), a principled and fully automated framework that explicitly models the \textit{dynamic contribution} of each training sample during the RFT process. Unlike previous heuristic-based methods, DIEM integrates data importance estimation directly into the policy optimization loop, making data utilization an adaptive component of learning rather than a fixed pretraining stage. At every optimization step, DIEM quantifies how much each sample contributes to the current policy improvement and dynamically reweights its influence in subsequent updates. This continuous adjustment transforms data selection from a one-time preprocessing step into a self-organizing, curriculum-like mechanism driven entirely by model dynamics.

Specifically, DIEM consists of two key components executed at every RFT iteration: (1) \textbf{Dynamic Data Importance Measuring} and (2) \textbf{Dynamic Data Reweighting}.
For \texttt{importance measuring}, our approach departs from heuristic criteria such as difficulty or entropy \citep{forgetting,entropy,zhang2025speedrlfastertrainingreasoning,gao2025pcl} and instead introduces a theoretically grounded, gradient-based estimator that directly evaluates each sample's marginal contribution to policy improvement. The estimator computes the alignment between the sample's gradient and the aggregated batch gradient, providing an efficient and mathematically guaranteed proxy for true importance.
For \texttt{data reweighting}, we formulate a constrained optimization problem that maximizes the batch's aggregate contribution while maintaining consistent gradient magnitude, thereby stabilizing the optimization trajectory and avoiding large step-size fluctuations. This joint mechanism converts data selection into an intrinsic part of RFT, enabling adaptive curriculum formation and stable credit assignment throughout training.

\begin{figure*}[tp]
\centering
\includegraphics[width=0.99999936\linewidth]{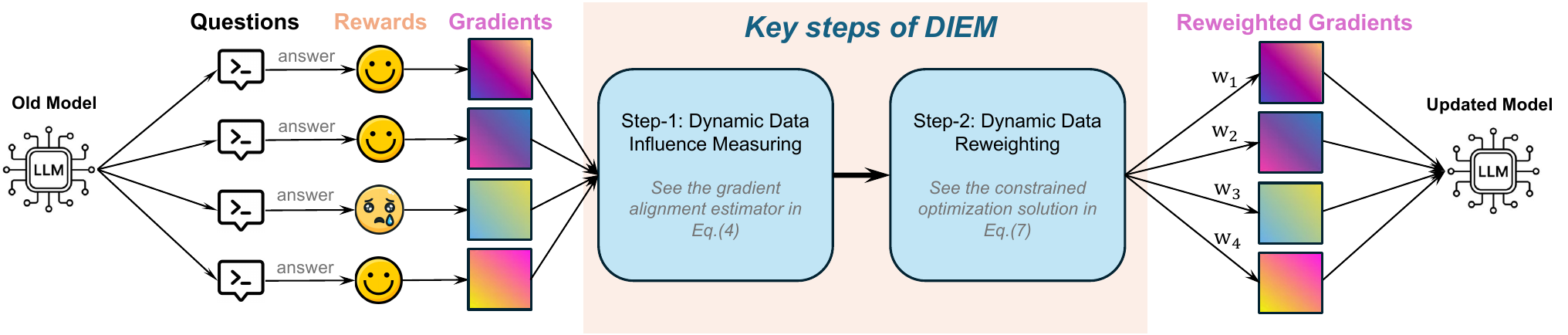}
\caption{\label{fig:pipeline} The pipeline of our proposed dynamic important example mining (DIEM).
Compared with the vanilla RFT pipeline, our DIEM introduces only two lightweight steps:
(1) \textbf{Dynamic Data Importance Measuring}, which evaluates the importance of each sample within the current batch using our proposed influence estimator in Eq.\ref{eq:gradient_alignment}; and
(2) \textbf{Dynamic Data Reweighting}, which reallocates sample weights based on their importance. The reweighting step is formulated as a constrained optimization problem that preserves the overall gradient magnitude while emphasizing the most important samples (defined in Eq.\ref{eq:reweighting_problem}, and we further provide an efficient closed-form solution in Eq.\ref{eq:final_closed_form_revised}).
Overall, DIEM is highly efficient, intuitive, and broadly compatible with existing RFT algorithms \citep{deepseek-math,REINFORCE++,schulman2017proximal,schulman2015trust}.
}
\end{figure*}

We extensively evaluate DIEM on large-scale multimodal reasoning benchmarks. Across all benchmarks, DIEM consistently outperforms strong static and dynamic baselines, providing a significant performance uplift to the base RFT algorithm of approximately $1\% \text{--} 6\%$, while introducing only a minimal additional $\mathbf{1.2\%}$ training overhead. Moreover, DIEM demonstrates broad compatibility with diverse RFT algorithms \citep{deepseek-math,schulman2015trust,schulman2017proximal,REINFORCE++}, confirming its general applicability. Furthermore, we demonstrate that this dynamic reweighting intrinsically induces a powerful, self-organizing curriculum learning paradigm (Sec.~\ref{sec: exp vis}).

We summarize our primary contributions as follows:
\begin{enumerate}
    \item We introduce a {novel, theoretically-grounded, fully automated, and highly efficient dynamic data selection framework} that significantly boosts the performance of RFT with negligible computational overhead.\vspace{0.1cm}
    \item We develop a dynamic importance indicator based on gradient-based alignment to precisely quantify the marginal contribution of individual samples, moving decisively beyond static or heuristic scoring methods. \vspace{0.1cm}
    \item DIEM consistently achieves the best performance with a minimal additional $\mathbf{1.2\%}$ training overhead. Furthermore, we demonstrate that this dynamic reweighting intrinsically induces a powerful, self-organizing curriculum learning paradigm (Sec.~\ref{sec: exp vis}).
\end{enumerate}

\section{Related Works}

\vspace{0.2cm}
\noindent\textbf{Reinforcement Fine-Tuning.}
The rapid surge in the performance of Large Models \citep{o1, deepseek-r1, team2025kimi} is largely attributable to Reinforcement Fine-Tuning (RFT). This optimization paradigm refines model behavior through reinforcement learning, leveraging reward signals to steer the model towards desired outputs.
The canonical RFT pipeline, particularly for aligning LLMs with human values, is Reinforcement Learning from Human Feedback (RLHF) \citep{christiano2023deepreinforcementlearninghuman,ouyang2022training}.
Recent advancements have diversified RFT away from the conventional RLHF with external RMs. For instance, in tasks such as complex reasoning and code generation, RFT can use verifiable rewards, where the reward is a direct, objective signal of correctness, such as a pass/fail from a unit test or a mathematical proof check \citep{deepseek-math, deepseek-r1}. DeepSeek-R1 \citep{deepseek-r1} is a prominent example, employing Group Relative Policy Optimization (GRPO) \citep{deepseek-math} with binary rewards to achieve state-of-the-art results.
Much of the contemporary RFT research focuses on algorithmic refinements to the standard PPO/RLHF process. For example, DAPO and GPG \citep{yu2025dapo, chu2025GPG} have sought to simplify GRPO for more robust empirical performance in certain settings.
Numerous studies are also applying Reinforcement Fine-Tuning (RFT) to multimodal reasoning \citep{huang2025vision,zhang2025r1vllearningreasonmultimodal,meng2025mmeureka,yang2025r1,chen2025sft,peng2025skywork,vision-r1-zhan,zhang2025scafgrposcaffoldedgrouprelative}. Examples of these contributions include the proposal of high-quality and diverse RFT datasets \citep{meng2025mmeureka}, the design of step-wise supervision signals like R1-VL \citep{zhang2025r1vllearningreasonmultimodal}, and the development of cost-efficient training pipelines \citep{chen2025r1v}.

\vspace{0.2cm}
\noindent\textbf{Data Selection in RFT.} Most contemporary data-centric strategies for RFT aim to identify high-value samples and are broadly divided into two categories: \textbf{1. Static Selection}: These methods \citep{li2025limr,wang20251datareinforcement} curate the training subset before RFT begins, relying on a one-time analysis of reward trends or variability. For instance, LIMR \citep{li2025limr} selects samples by analyzing changes in reward trends, indicating significant opportunities for policy improvement. Historical Variance Score HVS \citep{wang20251datareinforcement} prioritizes data exhibiting high reward variability, based on the intuition that instability signals high optimization impact.
Their critical flaw is the underlying assumption that a sample's importance remains constant throughout the entire non-stationary training process.
\textbf{2. Dynamic Selection}: Inspired by curriculum learning, these methods \citep{simplerl-zoo, ji2025difficulty, yu2025dapo} dynamically reorder samples, often prioritizing examples of intermediate difficulty. PCL \citep{gao2025pcl} trains an auxiliary value model to serve as an external difficulty evaluator. SPEED-RL \citep{zhang2025speedrlfastertrainingreasoning} uses the pass rate of samples as an importance indicator, often prioritizing medium-difficulty examples. However, because they rely on heuristic metrics external to the core policy optimization, they fail to accurately reflect the policy's true internal preference or the sample's genuine marginal contribution to the final policy update.

\section{Preliminaries} \label{ssec:preliminaries}

We now formalize the application of Reinforcement Learning (RL) to the finetuning of Large Models \citep{dubey2024llama3herdmodels, qwen2.5}, a process commonly referred to as Reinforcement Fine-Tuning (RFT) \citep{o1, deepseek-r1, deepscaler2025, deepseek-math, yuan2025vapo, team2025kimi}.
The training corpus is denoted as $\mathcal{Z} = \{z_i: (s_i, y_i)\}_{i=1}^N$, consisting of $N$ input-output pairs, where $s_i$ represents a prompt or query and $y_i$ is its solution. In domains such as mathematical reasoning, $y_i$ typically corresponds to a deterministic ground-truth answer.
The LLM acts as a stochastic policy $\pi_{\theta}$, parameterized by $\theta$. Within this framework, the input query $s$ serves as the observation (state), and the generated text $a$ corresponds to the chosen action. The quality of $a$ is evaluated via a scalar feedback signal-- the reward $r$. The objective of RFT is to iteratively optimize $\pi_\theta$ such that it generates high-fidelity, desirable responses \citep{schulman2017proximal, deepseek-math}. This is typically instantiated by assigning $r=1$ for exact matches with the target $y_i$ and $r=-1$ for incorrect or suboptimal generations.

\vspace{0.3cm}
\noindent\textbf{Vanilla Policy Optimization.}
The fundamental goal of RFT is realized by maximizing the expected empirical advantage over sampled state-action pairs $(s, a)$. This constitutes the simplest Policy Gradient optimization criterion and the corresponding policy gradient:
\begin{equation}
\begin{aligned}
    \mathcal{J}(\theta) &= \mathbb{E}_{(s, a)}\Big[ A(s,a) \Big],
\label{eq:po_loss_revised}\\
     \nabla_\theta \mathcal{J}(\theta) &= \mathbb{E}_{(s, a)}\Big[ A(s,a) \cdot \nabla_\theta \log \pi_\theta(a|s) \Big],
\end{aligned}
\end{equation}
where the advantage function is defined as $A(s, a) = r(s, a) - v$. This term quantifies the relative merit of executing action $a$ in state $s$ compared to a learned or predefined baseline value $v$. Optimization proceeds via iterative stochastic gradient ascent: $\theta_{t+1} = \theta_t + \eta_t \nabla_\theta J(\theta_t)$, where the step size is governed by the learning rate $\eta_t$.

\vspace{0.3cm}
\noindent\textbf{Beyond the Vanilla Policy Gradient.} To circumvent the inherent instability associated with simple gradient ascent, contemporary RFT architectures commonly integrate refined optimization schemes, like Proximal Policy Optimization (PPO) \citep{schulman2017proximal}, Quantile-Based Policy Optimization (QPO) \citep{jiang2022quantilebasedpolicyoptimizationreinforcement}, Group Policy Gradient (GPG) \citep{chu2025GPG}, Dynamic Sampling Policy Optimization \citep{yu2025dapo}.
Among them, Group Relative Policy Optimization (GRPO) \citep{deepseek-math, deepseek-r1} has emerged as one of the most influential techniques and now serves as the foundation for many RFT pipelines \citep{yu2025dapo, liu2025understanding, yuan2025vapo, zhang2025srpo}.
GRPO extends PPO by computing the advantage through intra-group comparisons of outputs generated for a single prompt $s$. Rather than training a separate value model to estimate the baseline $v$, GRPO samples a small set of responses $\{a_1, \dots, a_G\}$ from the old policy $\pi_{\theta_{\text{old}}}$. The GRPO objective retains the PPO-style clipping mechanism and is formulated as:
\begin{equation}
\small
\begin{aligned}
    &\mathcal{J}^{\text{GRPO}}(\theta) = \mathbb{E}_{(s, a_i)} \Big[ \frac{1}{G} \sum_{i=1}^G \min \Big( \frac{\pi_{\theta}(a_i|s)}{\pi_{\theta_{\text{old}}}(a_i|s)} A(s,a_i), \\
    &\text{clip} \Big( \frac{\pi_{\theta}(a_i|s)}{\pi_{\theta_{\text{old}}}(a_i|s)}, 1-\epsilon, 1+\epsilon \Big) A(s,a_i) \Big) - \beta\text{KL} \left( \pi_{\theta} \| \pi_{\text{ref}} \right) \Big],
\end{aligned}
\end{equation}
where, the advantage term $A(s,a_i)$ is calculated based on the standardized reward within the collected group, defined as $A(s,a_i) = \frac{r_i - \text{mean}(r_1, \dots, r_G)}{\text{std}(r_1, \dots, r_G)}$, where $r_i$ is the reward for output $a_i$. The clipping operator truncates excessively large or small relative advantage values, thereby ensuring a smoother and more stable optimization process. This standardization fundamentally bypasses the need for a separate value function, simplifying the training process. The $\text{KL}(\cdot)$ regularization term weighted by factor $\beta$ is included to penalize excessive divergence from a fixed reference model $\pi_{\text{ref}}$ (often the initial policy), ensuring a robust learning curve.

\section{Dynamic Important Example Mining}

The core idea of our Dynamic Important Example Mining (DIEM) is to assess the importance of the current batch of data at each step of the RFT process and adjust the weight of each sample accordingly.
Section~\ref{ssec:importance} will present our theoretically grounded strategy for measuring sample importance. Section~\ref{ssec:reweight} will describe the sample re-weighting strategy, and finally, Section~\ref{ssec:overview} will give an overview of the entire algorithm.

\subsection{Dynamic Data Importance Measuring}
\label{ssec:importance}

We begin by introducing a principled formulation for quantifying the contribution of each training sample to the model's learning dynamics. Unlike heuristic methods that rely on ad hoc proxies such as reward variance or difficulty, our approach provides a formal and theoretically justified estimate of each sample's true influence on policy improvement.

\vspace{0.3cm}
\noindent\textbf{Problem Definition.} Let $\mathcal{B}_t$ denote the training minibatch sampled at RFT step $t$, and let $z \in \mathcal{B}_t$ be an individual sample. We use $\mathcal{B}_t \setminus \{z\}$ to denote the batch with sample $z$ removed. Let $\theta_t$ be the policy parameters before the update at step $t$.
The performance metric we consider is the Total Batch Reward achieved by the policy after one gradient update.
We define the importance score $\mathcal{I}_t(z)$ of sample $z$ at step $t$ as the change in the total batch reward, evaluated on the original batch $\mathcal{B}_t$, that results from removing $z$ from the update calculation \citep{IF,hara2019data,TracIN,tan2025understanding, tan2025understandingdatainfluencedifferential}:
\begin{equation}
\mathcal{I}_t(z) =
\mathcal{J}\Big(\theta_{\mathcal{B}_t}^{\text{update}}, \mathcal{B}_t \Big)
-
\mathcal{J}\Big(\theta_{\mathcal{B}_t \setminus \{z\}}^{\text{update}}, \mathcal{B}_t \Big),
\label{eq:dynamic_importance_revised}
\end{equation}
where $\theta_{\mathcal{B}}^{\text{update}}$ is the policy $\pi_{\theta_{t}}$ after a single gradient step calculated on batch $\mathcal{B}$.
This metric offers a precise, non-heuristic measure of utility. A positive $\mathcal{I}_t(z)$ indicates that including sample $z$ in the update led to a better performing policy on the current batch's reward, suggesting $z$ is beneficial for that specific step. Conversely, a negative $\mathcal{I}_t(z)$ implies $z$'s inclusion caused a detrimental policy shift relative to the batch's total reward, suggesting it is harmful at this moment. Scores near zero denote a negligible marginal influence. This directional score thus serves as an interpretable indicator for dynamic sample reweighting.

\subsubsection{Efficient Estimation}

The direct computation of $\mathcal{I}_t(z)$ as defined in Eq.~\eqref{eq:dynamic_importance_revised} is intractable. Calculating $\mathcal{I}_t(z)$ for a batch of size $|\mathcal{B}_t|$ necessitates at least $|\mathcal{B}_t|$ times of full gradient updates and evaluation, a cost that renders the method unusable in high-throughput training environments. To address this severe computational bottleneck, we introduce an {\textit{Efficient First-order Gradient Alignment Estimator}} that serves as a highly effective, low-cost proxy for $\mathcal{I}_t(z)$. This estimator determines a sample's instantaneous value by quantifying the degree to which its policy gradient contributes to the collective batch update direction.

\vspace{0.3cm}
\noindent\textbf{Proposition 1. (Gradient Alignment Estimator)} \textit{We define the estimated importance score $\hat{\mathcal{I}}_t(z)$ for a sample $z \in \mathcal{B}_t$ at step $t$ using the inner product between its individual policy gradient and the total aggregate policy gradient of the batch:}
\begin{equation}
\hat{\mathcal{I}}_t(z) = \eta_t \Big\langle \mathcal{G}_z^{(t)} ,~ \mathcal{G}_{\mathcal{B}_t}^{(t) } \Big\rangle,
\label{eq:gradient_alignment}
\end{equation}
\textit{where the inner product $\langle \cdot, \cdot \rangle$ measures vector alignment. The terms are defined as follows: $\mathcal{G}_z^{(t)}$ represents the policy gradient vector contributed by the single sample $z=(s, y)$ to the update at step $t$. $\mathcal{G}_{\mathcal{B}_t}^{(t) }$ is the aggregate policy gradient across all samples in the current batch $\mathcal{B}_t$.}

\vspace{0.1cm}
This formulation is computationally lightweight, as all constituent gradient terms ($\mathcal{G}_z^{(t)}$ and $\mathcal{G}_{\mathcal{B}_t}^{(t) }$) are already calculated during the standard RFT backpropagation step.
The scalar score $\hat{\mathcal{I}}_t(z)$ offers an immediate and highly interpretable measure of sample utility: \textit{(1). Beneficial Alignment (High Positive Score):} A large positive value indicates that the sample's optimization direction ($\mathcal{G}_z^{(t)}$) is highly aligned with the overall direction of the batch ($\mathcal{G}_{\mathcal{B}_t}^{(t) }$). Such a sample is considered representative and high-utility, as its inclusion pulls the model in the same direction as the collective group, significantly contributing to efficient convergence. \textit{(2). Detrimental Divergence (Negative Score):} A negative value implies the sample's gradient is pointing away from the batch's aggregate direction. Including such a sample can slow down convergence or introduce noise, making it a low-priority or potentially harmful sample at that specific step.

\begin{algorithm}[tp]
\caption{Reinforcement Fine-Tuning with Dynamic Important Example Mining (\texttt{DIEM})}
\label{alg:reweighting}
\begin{algorithmic}[1]
\REQUIRE Full dataset $\mathcal{Z}$; Policy $\pi_{\theta_0}$; Total epochs $E$. \\
\FOR{epoch $e=1$ to $E$}
    \FOR{minibatch $\mathcal{B}_t \subset \mathcal{Z}$}
        \STATE \textcolor{blue}{//Dynamic Data Importance Measuring.}
        \STATE \vspace{0.1cm}Compute individual policy gradient $\mathcal{G}_{z}$ for each sample $z \in {\mathcal{B}_t}$ and aggregate them into a gradient matrix $\mathbf{G}$. \vspace{0.1cm}
        \STATE Calculate the sample-wise influence $\hat{\mathcal{I}}_t(z)$ via the gradient alignment estimator Eq.\eqref{eq:gradient_alignment} for $z \in \mathcal{B}_t$ and aggregate them into a importance vector $\mathbf{I} \in \mathbb{R}^N$. \vspace{0.1cm}
        \STATE \textcolor{blue}{//Dynamic Data Reweighting.}
        \STATE Compute Gram Matrix $\mathbf{P} = \mathbf{G} \mathbf{G}^\top$ and constant $C = \left\| \mathbf{1}^\top \mathbf{G} \right\|^2$ ($\mathbf{1}$ being the vector with all $1$ values). \vspace{0.1cm}
        \STATE Solve for unconstrained optimal weight vector $\mathbf{W}^*$ using Eq.\eqref{eq:final_closed_form_revised}. \vspace{0.1cm}
        \STATE {Post-Processing (Enforce Non-Negativity):} $\mathbf{W}^* \leftarrow \max(0, \mathbf{W}^*)$. \vspace{0.1cm}
        \STATE Calculate reweighted total gradient: $\mathbf{G}_{\text{weighted}} = \mathbf{W}^{*\top} \mathbf{G}$. \vspace{0.1cm}
        \STATE Update policy parameters: $\theta_{t+1} \leftarrow \theta_t + \eta_t \cdot \mathbf{G}_{\text{weighted}}$. \vspace{0.1cm}
    \ENDFOR
\ENDFOR
\STATE \textbf{Return} Final model parameters $\theta_E$
\end{algorithmic}
\end{algorithm}

\subsubsection{Theoretical Analysis}

We next provide a theoretical guarantee for the robustness of our gradient-based importance estimator, establishing its bounded approximation error under mild conditions.

\vspace{0.3cm}
\noindent\textbf{Proposition 2. (Error Bound)}
\textit{Assuming the policy's log-likelihood function is $\ell$-Lipschitz continuous and the advantage value is upper-bounded by $A_\text{max}$, the approximation error between the true Step-wise Importance $\mathcal{I}_t(z)$ and our estimated score $\hat{\mathcal{I}}_t(z)$ is bounded as follows:}
\begin{equation}
\Big|\mathcal{I}_t(z) - \hat{\mathcal{I}}_t(z)\Big| \leq \mathcal{O}\Big( \eta_t \ell^2 + 2\eta_t \ell A_{max}  \Big).
\label{eq:bound_revised}
\end{equation}

This proposition formalizes how the approximation error depends on key factors such as the learning rate ($\eta_t$) and the reward smoothness (Lipschitz constant $\ell$), yielding a bound that is even tighter than many established influence-function results in the simpler supervised learning setting \citep{hara2019data, wang2024capturing}.
Moreover, unlike conventional influence metrics \citep{IF,Studying_large_language,kwon2023datainf,group_rffect_analysis,Arnoldi}, our estimator does not rely on convexity assumptions or require the policy to be near a stationary point, making it naturally suited for the highly non-convex and non-stationary early-training regime of RFT.
Please refer to the appendix for the full derivation.

\subsection{Dynamic Data Reweighting}
\label{ssec:reweight}

\noindent\textbf{Problem Definition.}
Given the importance score for every sample in the minibatch, we formulate the dynamic data reweighting process as a constrained optimization problem. The objective is designed to maximize the batch's aggregate learning utility while actively preserving the stability of the optimization trajectory. Specifically, let $\mathbf{I} \in \mathbb{R}^N$ be the vector of importance scores for every sample in the minibatch $\mathcal{B}_t$, $\mathbf{W} \in \mathbb{R}^N$ be the sample weight vector, and $\mathbf{G} \in \mathbb{R}^{N \times D}$ be the matrix of individual policy gradients. The final reweighting problem is:
\begin{align}
\label{eq:reweighting_problem}
\max_{\mathbf{W}} \quad  \mathbf{I}^\top \mathbf{W} \quad \text{s.t.} \quad \left\| \mathbf{W}^\top \mathbf{G} \right\|^2 = \left\| \mathbf{1}^\top \mathbf{G} \right\|^2,
\end{align}
where the optimization objective is to maximize the weighted utility $\mathbf{I}^\top \mathbf{W}$, implicitly prioritizing high-impact samples. This constraint mandates that the $L_2$ norm of the reweighted aggregate gradient must equal that of the original unweighted gradient ($\mathbf{1}$ being the vector with all $1$ values), guaranteeing a stable and consistent update magnitude.

\vspace{0.3cm}
\noindent\textbf{Quasi-Closed-Form Solution.}
To maximize computational efficiency, we leverage the method of Lagrange multipliers to analytically solve the constrained optimization problem in Eq.\eqref{eq:reweighting_problem}. The full derivation, obtained by solving the stationary point of the Lagrangian, is detailed in the Supplementary Material.
We first define the necessary terms: $\mathbf{P} = \mathbf{G} \mathbf{G}^\top \in \mathbb{R}^{N \times N}$: the Gram matrix of the individual policy gradients, and $C = \left\| \mathbf{1}^\top \mathbf{G} \right\|^2$: the constant squared magnitude of the unweighted total gradient.
We present the solution below:
\begin{equation}
\mathbf{W}^* = \frac{\mathbf{P}^{-1} \mathbf{I}}{\sqrt{C}} \sqrt{\mathbf{I}^\top \mathbf{P}^{-1} \mathbf{I}}.
\label{eq:final_closed_form_revised}
\end{equation}
where the analytic solution is computationally efficient, requiring only a single inversion of the Gram matrix $\mathbf{P}$. Crucially, the dimension of $\mathbf{P}$ is defined by the small minibatch size $N$, rendering the computational overhead trivial compared to the overall duration of a single RFT step, which typically spans several minutes and even tens of minutes.

\vspace{0.3cm}
\noindent\textbf{Post-Processing and Policy Update.}
It is crucial to note that the analytically derived weight vector $\mathbf{W}^*$ by Eq.\eqref{eq:final_closed_form_revised} may contain negative values. A negative weight suggests that a sample is either truly detrimental to the current update or is misclassified as low-utility due to inherent estimation error.
To enforce the fundamental requirement that samples must contribute non-negatively to the policy strength, we apply a necessary post-processing step: we clip all negative weights to zero:
\[
\mathbf{W}^* \leftarrow \max\Big(0, \mathbf{W}^*\Big).
\]
This step effectively reintroduces the non-negativity constraint and ensures that the final weight vector only consists of components that contribute constructively to the optimization direction.
Subsequently, this post-processed weight vector $\mathbf{W}^*$ yields the reweighted gradient $\mathbf{G}_{\text{weighted}} = \mathbf{W}^{*\top} \mathbf{G}$. This weighted gradient is then used for the RFT policy update: $\theta_{t+1} = \theta_t + \eta_t \cdot \mathbf{G}_{\text{weighted}}$. This dynamic mechanism establishes a self-organizing curriculum that efficiently leverages high-impact samples without compromising the core RFT optimization stability.

\subsection{Algorithm Overview}
\label{ssec:overview}

Our proposed approach, which we term Reinforcement Fine-Tuning with Dynamic Important Example Mining (\texttt{DIEM}), integrates the importance estimation and reweighting mechanism directly into the standard optimization loop of any policy gradient RFT method (e.g., PPO or GRPO). Unlike static data selection pipelines that require pre-training and subset formation, our method continuously modulates sample contributions throughout the entire training process, as detailed in Algorithm~\ref{alg:reweighting}.
For every optimization step $t$ using a minibatch $\mathcal{B}_t$, the algorithm executes as follows:

\begin{enumerate}
    \item \textbf{Gradient and Importance Calculation (Lines 4-5):} The individual policy gradients ($\mathcal{G}_z^{(t)}$) for all samples in $\mathcal{B}_t$ are computed. This is used to derive the total gradient matrix $\mathbf{G}$. The Importance Score vector $\mathbf{I}$ is then calculated by taking the inner product between each gradient and the aggregate batch gradient by Eq.~\eqref{eq:gradient_alignment}. \vspace{0.1cm}
    \item \textbf{Solving for Optimal Weights (Lines 7-8):} The algorithm solves the original constrained optimization problem to find the optimal weight vector $\mathbf{W}^*$ using the quasi-closed-form solution by Eq.~\eqref{eq:final_closed_form_revised}. This step determines the weighting that maximizes sample utility while satisfying the gradient magnitude preservation constraint. \vspace{0.1cm}
    \item \textbf{Post-Processing and Update (Lines 9-11):} The analytically derived $\mathbf{W}^*$ is post-processed by clipping any resulting negative weights to zero. This enforces the non-negativity principle, ensuring only constructive information contributes to the update. The final $\mathbf{W}^*$ is then used to calculate the reweighted gradient $\mathbf{G}_{\text{weighted}}$.
\end{enumerate}

\section{Experiments}

In Sec.~\ref{sec: exp main llm} and Sec.~\ref{sec: exp main}, we report the main results across visual and mathematical reasoning benchmarks. Finally, we provide two dedicated analyses: an ablation study in Sec.~\ref{sec: exp ablation} and a visualization in Sec.~\ref{sec: exp vis}.

\subsection{LLM Results}
\label{sec: exp main llm}

All algorithms are implemented via the veRL framework using open-source Qwen series models, and trained on a cluster of 16 NVIDIA H200 GPUs. We compare our approach against GRPO \citep{deepseek-math}, LIMR \citep{li2025limr}, and HVS \citep{wang20251datareinforcement} baselines. For the GRPO \citep{deepseek-math} baseline, we disable the KL penalty and entropy bonus, utilizing a clipping ratio of 0.2. The training configurations include a prompt batch size of 64, 8 rollouts per prompt, a mini-batch size of 32, and a micro-batch size of 8 (reduced to 4 for 7B/8B models). The maximum prompt and response lengths are set to 1,024 and 2,048 tokens, respectively. We optimize the models with a constant learning rate of 1e-6 and no warmup. The training data comprises 14,973 math problems (7,500 from MATH and 7,473 from dapo-math). 

\begin{table}[htbp]
  \centering
  \scriptsize
  \caption{Performance comparison on different benchmarks across various models.}
  \label{tab:model_comparison_llm}
  \begin{tabular}{llrrrrrr} 
    \toprule
    \textbf{Model} & \textbf{Method} & \textbf{MATH-500} & \textbf{Gaokao23en} & \textbf{AMC-23} & \textbf{AIME24} & \textbf{AIME25} & \textbf{Avg} \\
    \midrule
    \multirow{4}{*}{Qwen-3-1.7B} 
    & GRPO \citep{deepseek-math}  & 59.9 & 48.4 & 37.2 & 7.9 & 3.4 & 31.36 \\
    & HVS \citep{wang20251datareinforcement} & 60.2 & 48.9 & 37.8 & 7.9 & 3.8 & 31.72 \\
    & LIMR \citep{li2025limr} & 60.6 & 49.5 & 38.4 & 8.2 & 4.3 & 32.20 \\
    & DIEM & {61.3} & {51.1} & {39.7} & {7.9} & {5.5} & {33.10} \\
    \midrule
    \multirow{4}{*}{Qwen2.5-3B} 
    & GRPO \citep{deepseek-math} & 52.2 & 44.8 & 35.9 & 5.0 & 1.2 & 27.82 \\
    & HVS \citep{wang20251datareinforcement} & 52.9 & 45.1 & 36.5 & 5.3 & 1.4 & 28.24 \\
    & LIMR \citep{li2025limr} & 53.8 & 45.7 & 37.6 & 5.5 & 1.3 & 28.78 \\
    & DIEM & {55.8} & {46.8} & {41.9} & {5.5} & {1.6} & {30.32} \\
    \midrule
    \multirow{4}{*}{Qwen3-4B} 
    & GRPO \citep{deepseek-math} & 59.2 & 42.5 & 58.5 & 13.8 & 12.5 & 37.30 \\
    & HVS \citep{wang20251datareinforcement} & 60.1 & 44.6 & 58.9 & 14.2 & 12.9 & 38.14 \\
    & LIMR \citep{li2025limr} & 61.5 & 47.0 & 60.1 & 14.9 & 13.6 & 39.42 \\
    & DIEM & {65.9} & {52.0} & {55.0} & {16.2} & {14.2} & {40.66} \\
    \midrule
    \multirow{4}{*}{Qwen2.5-7B} 
    & GRPO  \citep{deepseek-math} & 54.9 & 43.6 & 53.5 & 12.5 & 5.5 & 34.00 \\
    & HVS \citep{wang20251datareinforcement} & 55.5 & 43.8 & 53.9 & 12.7 & 6.0 & 34.38 \\
    & LIMR \citep{li2025limr} & 56.2 & 44.1 & 54.5 & 12.9 & 7.2 & 34.98 \\
    & DIEM & {58.8} & {44.3} & {51.2} & {13.3} & {10.8} & {35.68} \\
    \bottomrule
  \end{tabular}
\end{table}

Table~\ref{tab:model_comparison_llm} compares DIEM against GRPO and two intermediate baselines (HVS and LIMR) across four model families. We focus on the DIEM vs.\ GRPO comparison, as GRPO serves as the standard RLHF baseline for reasoning. DIEM consistently outperforms GRPO on the average metric across all four models, with absolute gains of +1.74 on Qwen-3-1.7B (31.36 $\to$ 33.10), +2.50 on Qwen2.5-3B (27.82 $\to$ 30.32), +3.36 on Qwen3-4B (37.30 $\to$ 40.66), and +1.68 on Qwen2.5-7B (34.00 $\to$ 35.68). The improvements are particularly substantial on the high-difficulty competition benchmarks: on AIME25, DIEM improves over GRPO by +2.1 on Qwen-3-1.7B (3.4 $\to$ 5.5) and +5.3 on Qwen2.5-7B (5.5 $\to$ 10.8), representing relative gains of 62\% and 96\% respectively. The one exception is AMC-23 on Qwen3-4B, where DIEM scores 55.0 compared to GRPO's 58.5; this appears to be an outlier specific to this model--benchmark pair, as DIEM improves AMC-23 on all other models. HVS and LIMR, which apply heuristic and loss-based data selection strategies respectively, show marginal and inconsistent improvements over GRPO but fall noticeably short of DIEM, suggesting that the gains from DIEM stem from a fundamentally different data selection mechanism rather than incremental refinements of existing approaches.

\begin{table*}[thtp]
  \scriptsize
    \caption{\label{tb:cls} To examine the effectiveness of our DIEM algorithm, we compare our method against other data selection algorithms, and also famous large models. The selected benchmarks include both domain-specific and general-purpose tasks.} 
\setlength{\tabcolsep}{3.1pt}
\centering 
\begin{tabular}{l ccc ccccccccccccccc}
\toprule
            \textbf{Model}  & \textbf{MathVista} & \textbf{MathVerse} & \textbf{MathVision}    & \textbf{MMStar} & \textbf{MMMU} & \textbf{AI2D}   & \textbf{Avg} \\
            \midrule 
            GPT-4o \citep{Hurst2024GPT4o} & 63.8 & 50.8 & 30.4 & 65.1 & 70.7 & 84.9 & 60.9 \\
            Claude3.7-Sonnet \citep{Anthropic2024Claude35} &  66.8 & 52.0 & 41.3 & -- & 71.8   & -- & -- \\
            GPT-5-nano \citep{OpenAI2025GPT5} & 73.1    &66.6 &59.7 & -- & 72.6  & -- & -- \\
            \midrule
            \multicolumn{8}{c}{\small \em Base Model: Qwen2.5-VL-7B } \\ \midrule
            Qwen2.5-VL-7B \citep{bai2025qwen25vltechnicalreport} & 68.2 & 49.2 & 25.1   & 63.9 & 58.6  & 83.9 & 58.2\\
            Vanilla RFT \citep{deepseek-math} &  74.1  & 51.7  & 27.6    & 65.6  & 57.1 & 78.4  &59.1  \\
            LIMR  \citep{li2025limr}  &71.5 &52.4  & 25.7  &64.2 & 55.9  & 80.0  &58.3 \\
            HVS  \citep{wang20251datareinforcement} &  73.3 & 50.1 &26.2   & 66.4  & 58.5   & 82.7  &59.5 \\
            SPEED-RL \citep{zhang2025speedrlfastertrainingreasoning} & 74.9  & 47.1  & 27.5    & 66.7  & 59.2 & 84.7  &60.0 \\
            PCL  \citep{gao2025pcl}  &  70.5 & 51.3 &24.9   & 64.0  &  \textbf{59.3}  & 82.8  &58.8  \\
            \midrule
            \rowcolor{exp_table_blue} DIEM (ours)  &  \textbf{76.9}  &  \textbf{53.0} &  \textbf{28.7}   &  \textbf{67.9} & 59.2 &  \textbf{85.0} & \textbf{61.8}  \\
            \midrule
            \multicolumn{8}{c}{\small \em Base Model: Qwen2.5-VL-32B } \\ \midrule
            Qwen2.5-VL-32B \citep{bai2025qwen25vltechnicalreport} & 74.2 & 49.4 & 38.0 & 68.8 & 69.2  & 83.2 & 63.8 \\
            Vanilla RFT \citep{deepseek-math} & 75.6 & 52.7  & 39.1   & 69.2  & 69.0   &  84.3 & 64.9 \\
            LIMR  \citep{li2025limr}   & 74.2  & 54.1  & 39.8    & 67.5  & 69.9  &  83.9 & 64.9  \\
            HVS  \citep{wang20251datareinforcement} & 76.1 & 53.0  & 35.4     & 68.7  & 68.5  &  81.8 & 63.9 \\
            SPEED-RL  \citep{zhang2025speedrlfastertrainingreasoning} & 75.1  & 53.9  & 40.1    & 69.8  & 70.6 &  83.8 & 65.6  \\
            PCL   \citep{gao2025pcl}  & 73.6 & 54.1  & 39.7    & 70.5  & 68.8  &  85.1 & 65.3  \\
            \midrule
            \rowcolor{exp_table_blue} DIEM (ours) & \textbf{76.9}  & \textbf{58.0} & \textbf{40.3}    & \textbf{71.2} & \textbf{71.9} & \textbf{85.2} & \textbf{67.3} \\
            \bottomrule
\end{tabular}
\end{table*}

\subsection{VLM Results}
\label{sec: exp main}

We conducted our experiments using two scales of the Qwen2.5-VL model family \citep{bai2025qwen25vltechnicalreport}: the 7B and 32B variants. For training data, we randomly sampled 52K multimodal entries from the MM-Eureka corpus \citep{meng2025mmeureka}. The 7B model was trained on 16 NVIDIA A100 GPUs, and the 32B model on 32 GPUs. The reinforcement learning phase utilized a global batch size of 128 and a learning rate of $1 \times 10^{-6}$. We evaluate the performance of our models across a comprehensive suite of challenging multimodal benchmarks, covering both general reasoning and specialized mathematical problem-solving. Our evaluation set includes MathVista \citep{lu2023mathvista}, MathVerse \citep{zhang2024mathverse}, MathVision \citep{wang2024measuring}, MMStar \citep{chen2024we}, MMMU \citep{yue2024mmmu},  AI2D \citep{kembhavi2016diagram}.
We choose several baselines as competitors to our approach, including static selection methods, including HVS \citep{wang20251datareinforcement} and LIMR \citep{li2025limr}, and dynamic methods, including PCL \citep{gao2025pcl} and SPEED-RL \citep{zhang2025speedrlfastertrainingreasoning}. In addition, we also chose random selection as the most classic baseline.
The main experimental results, comparing our proposed \textbf{DIEM} algorithm against various state-of-the-art data selection methods and strong commercial models, are summarized in Table~\ref{tb:cls}.

Across the $7\text{B}$ model family, our proposed \textbf{DIEM} algorithm achieves a superior average performance of $\mathbf{61.8\%}$ across all six evaluated benchmarks, demonstrating its effectiveness in dynamic sample selection during RFT. This result represents a significant gain of $3.6$ percentage points over the Vanilla RFT baseline ($59.1\%$) and $\mathbf{1.8}$ percentage points over the best-performing existing data selection method, SPEED-RL ($60.0\%$) \citep{zhang2025speedrlfastertrainingreasoning}. Specifically, DIEM secures the highest scores in five out of six benchmarks, including MathVista ($76.9\%$), MathVerse ($53.0\%$), MathVision ($28.7\%$), MMStar ($67.9\%$), and AI2D ($85.0\%$). Notably, DIEM's performance on the average benchmark score ($\mathbf{61.8\%}$) also \textit{surpasses} the average score of the powerful commercial model GPT-4o ($60.9\%$), highlighting the efficiency and power of our data reweighting mechanism even on a much smaller base model. The performance advantage of DIEM is consistently maintained and further amplified when applied to the larger Qwen2.5-VL-32B base model. DIEM achieves a leading average score of $\mathbf{67.3\%}$, which is an improvement of $\mathbf{2.4}$ percentage points over the Vanilla RFT baseline ($64.9\%$) and a gain of $\mathbf{1.7}$ percentage points over the previous best dynamic baseline, SPEED-RL ($65.6\%$). For the $32\text{B}$ model, DIEM demonstrates \textbf{dominance across all six individual benchmarks}, achieving the highest score in every category. This comprehensive superiority underscores the ability of DIEM to consistently identify and prioritize high-value training samples, leading to more robust and effective policy optimization across diverse multimodal reasoning tasks, especially within higher-capacity models.

\begin{figure}[tp]
\centering
\includegraphics[width=0.499936\linewidth]{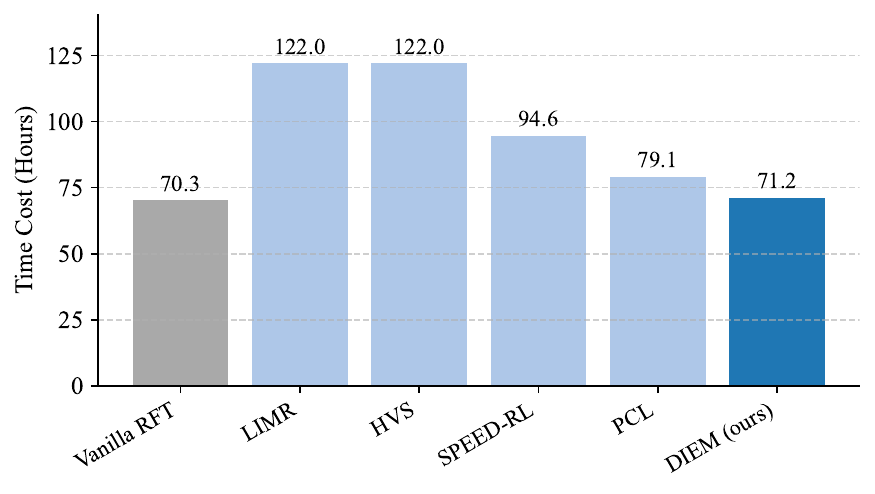}
\caption{\label{fig:speed}
The speed test of various data selection methods when combined with Vanilla RFT (GRPO) \citep{deepseek-math}.}
\end{figure}

\subsection{Ablation Study}
\label{sec: exp ablation}

Table~\ref{tab:ablation} presents an ablation study validating the contribution of DIEM's two core components against the full model's $\mathbf{58.0}$ performance. \textbf{The first group of studies} indicates that substituting our dynamic influence score with alternative metrics resulted in significant performance drops. Random value replacement or using the raw Pass@k score \citep{zhang2025speedrlfastertrainingreasoning} caused deficits of $\mathbf{5.0}$ (to 53.0) and $\mathbf{4.8}$ points (to 53.2), respectively. The largest drop was $\mathbf{5.9}$ points (to 52.1) when using the raw PCL Difficulty score \citep{gao2025pcl}. While translating these alternative scores to the distance to the median offered some improvement (e.g., Pass@k distance gained $\mathbf{1.7}$ points, reaching 54.9), none approached our method, confirming the superiority of our dynamic scoring approach. \textbf{The second group of studies} indicates that our reweighting function is essential. Removing this step entirely (NULL-operation) dropped performance by $\mathbf{2.6}$ points (to 55.4). Replacing it with a standard Softmax normalization resulted in a $\mathbf{1.6}$ point deficit (to 56.4). This confirms that our specialized reweighting function is critical for maximizing benefit from the influence scores. In summary, both the dynamic data influence score and the custom data reweighting step are indispensable components, jointly contributing to the superior $\mathbf{58.0}$ performance.

\begin{table}[tp]
    \centering
    \caption{Ablation study on MathVerse \citep{zhang2024mathverse} with Qwen-2.5-VL-32B \citep{bai2025qwen25vltechnicalreport} as the base model. Results on more benchmarks could be seen in the supplementary materials.
    }
    \resizebox{0.45999\linewidth}{!}{
    \begin{tabular}{l|cccccc}
    \toprule
    \rowcolor{exp_table_blue} Replace the dynamic data influence score with: &Performance \\
    \midrule
    \rowcolor{mygray} {\em No Replacement (Ours)} &58.0 \\
    Random value &53.0 \\
    Pass@k score \citep{zhang2025speedrlfastertrainingreasoning}  &53.2 \\
    Pass@k score to median distance  &54.9 \\
    Difficulty score \citep{gao2025pcl}   &52.1 \\
    Difficulty score to median distance    &53.6 \\
    \midrule
    \rowcolor{exp_table_blue} Replace the data reweighting step with: &Performance \\
    \midrule
    \rowcolor{mygray} {\em No Replacement (Ours)} &58.0 \\
    NULL-operation &55.4 \\
    Softmax normalization  &56.4 \\
    \bottomrule
    \end{tabular}
    }
    \label{tab:ablation}
\end{table}

\subsection{Speed Test}
DIEM offers a significant speed advantage, see Fig.~\ref{fig:speed}. Unlike traditional, static data selection models that demand time-consuming, resource-intensive training of a separate surrogate model on all data, our dynamic DIEM method eliminates this requirement. The speed test results confirm this efficiency: integrating DIEM with the baseline Vanilla RFT (GRPO) framework introduces only a minimal overhead. The baseline takes $70.3$ hours, while the combined DIEM system requires just $71.2$ hours, an increase of only $0.9$ hours (about $1.28\%$). This marginal increase is significantly lower than all other state-of-the-art methods tested, such as PCL ($79.1$ hours), SPEED-RL ($94.6$ hours), and especially LIMR and HVS (both $122.0$ hours). This superior efficiency stems from DIEM's design: it avoids separate inference or dedicated training steps for a surrogate model or data selector. Instead, DIEM operates solely by leveraging and reusing the gradient information already computed during the original RFT algorithm's standard update cycle.

\subsection{Training Process Visualization}
\label{sec: exp vis}

\begin{figure}[http]
\centering
\includegraphics[width=0.51999936\linewidth]{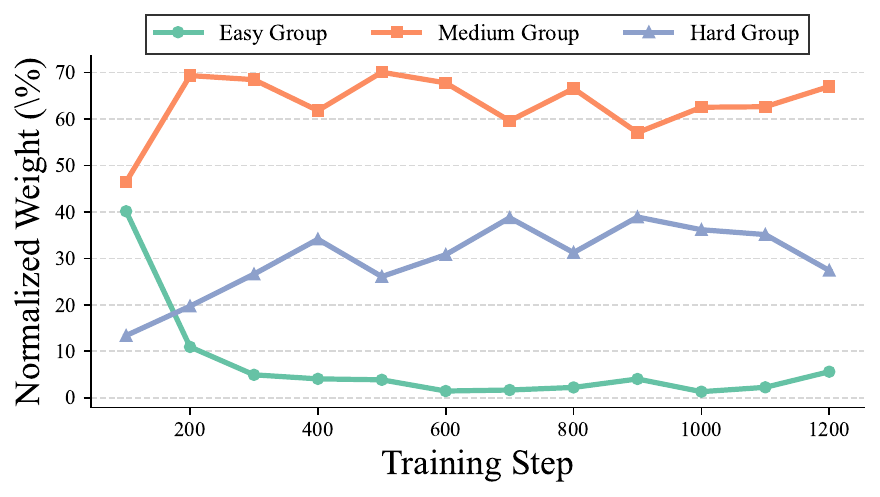}
\caption{\label{fig:kecheng}
Normalized weight trajectories of different difficulty samples (groups) during DIEM Training.}
\end{figure}

To quantify this evolution, we use Pass@k as a proxy for sample difficulty and group examples into Hard, Medium, and Easy categories (lower Pass@k implies higher difficulty). We then track the DIEM-assigned relative weights of these groups over training, as shown in Fig.~\ref{fig:kecheng}.
The results reveal that DIEM naturally introduces a smooth curriculum learning trend. During the initial training stages, both the Easy and Medium difficulty groups are assigned relatively high importance weights. However, as training progresses, the weight assigned to the Easy samples rapidly declines, indicating that the model has saturated its learning potential from these examples. Simultaneously, the weight trajectory of the Hard group shows a sustained increase, though accompanied by oscillations, as the model shifts its focus to more challenging samples.
This easy-to-hard progression emerges organically, in sharp contrast to prior methods \citep{gao2025pcl,zhang2025speedrlfastertrainingreasoning} that rely on rigid, hand-designed heuristics to adjust sample priority. Such heuristic approaches are often brittle and fail to reflect the model's evolving state, leading to less efficient learning.

\section{Conclusion}

We introduced the Dynamic Important Example Mining (DIEM) framework for enhancing Reinforcement Fine-Tuning (RFT) of large models. DIEM moves beyond fixed-value assumptions by using a theoretically-grounded, gradient-based importance estimator that efficiently quantifies a sample's marginal contribution to policy improvement. This innovation integrates data selection as a dynamic, self-organized component of the RFT process, forming a model-driven curriculum. Furthermore, our constrained optimization procedure for adaptive batch reweighting maximizes the collective contribution of selected samples while ensuring stable policy updates. Extensive experiments demonstrate both the superior performance and high efficiency of our method.

\bibliography{main}
\bibliographystyle{iclr2026_conference}

\end{document}